\documentclass[conference]{IEEEtran}
\IEEEoverridecommandlockouts
\usepackage{epsfig,makeidx,color}
\usepackage{amsmath,amssymb,bbm}
\usepackage{cite,graphicx,lipsum}
\usepackage{enumerate}
\usepackage{multirow,tabularx}
\usepackage[switch,pagewise]{lineno}
\usepackage{hyperref}
\hypersetup{
        colorlinks = true,
        citecolor=blue,
}
\def\cX{{\cal X}}

\def\rT{{\rm T}}

\def\uR{{\mathbb R}}

\def\uE{{\mathbb E}}

\def\be{ \begin{equation} }
\def\ee{ \end{equation} }
\def\bea{ \begin{eqnarray} }
\def\eea{ \end{eqnarray} }

\def\bx{{\bf x}}
\def\by{{\bf y}}

\def\bz{{\bf z}}

\def\bee{{\bf e}}

\def\bv{{\bf v}}

\def\bC{{\bf C}}

\def\bE{{\bf E}}

\def\bI{{\bf I}}

\def\bX{{\bf X}}

\def\b0{{\bf 0}}

\def\bSigma{{\bf \Sigma}}

\def\bLambda{{\bf \Lambda}}

\def\cA{{\cal A}}

\def\cZ{{\cal Z}}

\def\cN{{\cal N}}

\ifCLASSOPTIONonecolumn
\else

\fi
\usepackage{graphicx}
\usepackage{booktabs}
\usepackage{graphicx}
\usepackage{subcaption}
\begin{document}

% \title{A Subspace-based Approach to Anomaly Detection via Variational Autoencoder}
\title{A Subspace Projection Approach to Autoencoder-based Anomaly Detection}

% \title{A Subspace Projection Approach to Autoencoder-based Anomaly Detection
% \thanks{This research was supported
% by the Australian Government through the Australian Research
% Council's Discovery Projects funding scheme (DP200100391).
% }
% }

% \author{%
% Jinho Choi\affmark[1], Jihong Park\affmark[1],  Abhinav Japesh\affmark[2], and Adarsh Yadav\affmark[2] \\
% \affaddr{\affmark[1]School of Information Technology, Deakin University,  Australia }\\
% \affaddr{\affmark[2]IIT Kharagpur, %West Bengal 721302, 
% India}\\
% }

\author{$^\dagger$Jinho Choi, $^\dagger$Jihong Park, Abhinav Japesh, and Adarsh\\ 
\thanks{J. Choi and J. Park are with
the School of Information Technology,
Deakin University, Geelong, VIC 3220, Australia
(e-mail: \{jinho.choi, jihong.park\}@deakin.edu.au). A. Japesh and A. Yadav (equal contributions) are with IIT Kharagpur, West Bengal 721302, India (email: \{akjapesh, adygkp1130\}@gmail.com). This research was supported
by the Australian Government through the Australian Research
Council's Discovery Projects funding scheme (DP200100391).}}

\maketitle

\begin{abstract}
% In this paper, we study an approach to anomaly detection
% using variational autoencoder (VAE), which is
% one of popular unsupervised learning methods.  
% From the information bottleneck (IB) point of view,
% VAE is to approximate an input, $\bx$, through a minimal sufficient latent representation, denoted by $\bz$, that is obtained by imposing a bottleneck 
% of information flow. This property of VAE 
% can play a crucial role in anomaly detection as anomalies may not be reconstructed by their deemed minimal sufficient latent representations. 
Autoencoder (AE) is a neural network (NN) architecture that is trained to reconstruct an input at its output. By measuring the reconstruction errors of new input samples, AE can detect anomalous samples deviated from the trained data distribution. The key to success is to achieve high-fidelity reconstruction (HFR) while restricting AE's capability of generalization beyond training data, which should be balanced commonly via iterative re-training. Alternatively, we propose a novel framework of AE-based anomaly detection, coined \emph{HFR-AE}, by projecting new inputs into a subspace wherein the trained AE achieves HFR, thereby increasing the gap between normal and anomalous sample reconstruction errors. Simulation results corroborate that HFR-AE improves the area under receiver operating characteristic curve (AUROC) under different AE architectures and settings by up to $13.4$\% compared to Vanilla AE-based anomaly detection.
% In particular, the reconstruction error, which is the difference between the input and output of trained VAE, can be used as a test statistics for anomaly detection. Since VAE does not aim to exactly reproduce the input, it has a certain level of inherent reconstruction error, which may limit the performance of anomaly detection. To mitigate this problem, we propose a subspace method that is based on the subspace where the projected signal can be reliably reconstructed through VAE. Thus, any anomalous input to the VAE may likely have a significant projected reconstruction error on this subspace, while a normal input has a negligible projected reconstruction error.
% {\color{red} Compared to the conventional anomaly detection approach based on VAE, the proposed approach has a significant performance improvement. For example, with MNIST datasets, about 30\% performance improvement is observed in terms of the area under receiver operating characteristic curve (AUROC).
% }
\end{abstract}

\section{Introduction}

Anomaly  detection is a task to detect samples that differ from most of the data or deviate from some form of normality, and has a wide range of applications ranging from detecting fraud and intrusion to fault diagnosis \cite{Hodge04, Chandola09}. Various approaches to anomaly detection have been studied, and some of classical approaches are well summarized in \cite{Chandola09}. Recently, deep learning has been widely applied to anomaly detection \cite{Goel20,Pang21}, in which autoencoder (AE) architectures play an important role. An AE is a neural network (NN) that aims to reconstruct its input at the output. As an NN, a trained AE is inherently biased to its training data, so often fails to reconstruct outliers generated from a shifted distribution from that of training data, i.e., out-of-distribution (OOD) data. By turning such vulnerability to OOD data for reconstruction into advantages, the trained AE can be utilized for detecting anomalous data associated with high reconstruction errors \cite{An15}.

% out-of-sample (OOS) errors
% out-of-distribution (OOD) errors

% high fidelity reconstruction under limited training samples 
% overfitting the distribution of training data

% To exploit AE as a anomaly detector, AE should be intentionally overfitting training data.

% intentionally is intentionally overfit to 

% With AE architectures, both anomaly detection and reconstruction tasks have a commonality in that they aim to 

% out-of-sample error
% out-of-distribution error 

% intentionally overfitting to training data distribution

The success of AE based anomaly detection rests on achieving high-fidelity reconstruction (HFR) while restricting generalization capability. To this end, existing methods focus mostly on imposing and controlling an information bottleneck (IB) \cite{tishby2015deep}, so as to sift out spurious information and to learn only meaningful features. While the vanilla AE coarsely adjusts the discrete dimension of its hidden-layer activation (i.e., a latent variable), variational AE (VAE) enforces Gaussian-distributed latent variables \cite{Kingma14}, enabling its variant $\beta$-VAE to flexibly fine-tune IB \cite{higgins2017beta}. Vector-quantized VAE (VQ-VAE) additionally quantizes the latent variables of VAE \cite{Oord17}, provisioning qunderizer's codebook size as another dimension of fine-tuning IB. Notwithstanding, finding an optimal IB entails multiple rounds of re-training. Furthermore, optimal IBs for HFR and restricted generalization may not always be consistent, particularly when there is only a subtle difference between normal and anomalous samples (e.g., a single dataset divided into normal and anomalous classes).

\begin{figure}
    \centering
    \includegraphics[width=\columnwidth]{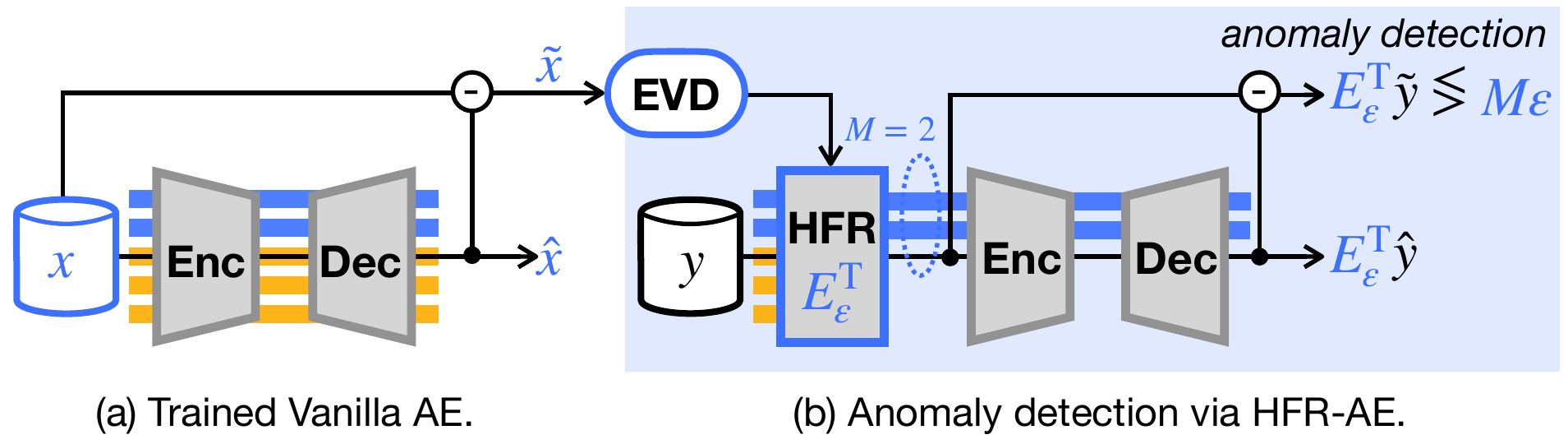}
    \caption{\small  A schematic illustration of anomaly detection using an autoencoder (AE) projecting an input $y$ into the high-fidelity reconstruction (HFR) subspace of training data $x$.}
    \label{fig:HFRAE} \vspace{-10pt}
\end{figure}

% apple (red and green) and grape (red green)

% different domain: 
% inspired from wireless 
% `channels'

Alternatively, in this article we propose an HFR-subspace projection approach to AE for anomaly detection, as Fig.~\ref{fig:HFRAE} illustrates. The resultant \emph{HFR-AE} framework is NN architecture-agnostic and free from re-training. Inspired from wireless communication, the key new element is to treat a trained AE between its input and output as multiple-input multiple-output (MIMO) channels \cite{TseBook05}, and divide them into two groups: HFR and low-fidelity reconstruction (LFR) channels resulting in low and high reconstruction errors, respectively. Then, a new input is projected onto the HFR channel subspace before feeding into the AE. Such projection increases the reconstruction error gaps between normal and anomalous samples, thereby helping distinguish them even when there is only a subtle difference in their original sample space. Furthermore, the key design parameter of HFR-AE is the threshold separating HFR and LFR channels, which can be optimized by simply feeding multiple samples without re-training the AE. 

Simulation results with CIFAR-10 dataset show that HFR-AE improves the area under receiver operating characteristic (AUROC) for anomaly detection under different AE architectures (i.e., Vanilla AE, VAE, and VQ-VAE) and different levels of IB (i.e., latent dimension) by up to  $13.4$\%. It is worth noting that AE has often been utilized for modeling a communication system in which the channel only implies the encoder-decoder connection \cite{nemati2022all}, whereas HFR-AE treats the entire AE as a channel. Subspace-based decomposition on an NN has also been done over the input weight of a decoder (or equivalently a generator) \cite{shen2021closed}, while HFR-AE applies the decomposition to the output of a decoder.

% 4.27 AE
% 7.25 VAE
% 4.36 VQ-VAE

% 13.4 (K=2)

% SeFa, decomposing decoder (generator)'s first-layer weights in the latent space (in the domain of hidden-layer activations), i.e., first-layer input-output channel; whereas we focus on the first-layer intput and last-layer output 

% AE based anomaly detection accuracy rests on the reconstruction fidelity of the AE. Indeed, a poor AE mapping all inputs into random data (e.g., blurry images) cannot distinguish between anomalous and normal data. Therefore, recent studies in this literature have been focusing on improving the architectures and training methods of AE [REFs]. Controlling the information bottleneck of the AE is central to most of these methods. When feeding forward data through an AE, the input data are intentionally compressed at the encoding layers before reconstruction by the decoding layers.

\section{Anomaly Detection via VAE}

Throughout this paper, we consider VAE as our baseline AE architecture. In this section, we briefly introduce VAE and its application to anomaly detection.

% In this section, we briefly discuss the anomaly detection
% approach based on VAE.

\subsection{VAE Architecture and Operations}

VAE is a deep Bayesian network which uses an NN to relate variables via dimensionality reduction and hence can be applied to different distribution families \cite{Kingma14}. The encoder-decoder architecture chooses the best scheme to relate a latent sample $\bz \in \cZ$ and a data point $\bx \in \cX$,
where $\cZ$ and $\cX$ are the latent space and data space, respectively.
Instead of encoding each data point to a latent sample, VAE encodes it as a distribution over the latent space which can be used for a generative purpose as well. 

Suppose that a dataset
$\bX =\{\bx(i), i = 1,\ldots, N\}$ is given, where $\bx(i) \in \cX$ represents an iid sample and $N$ is the number of samples.
A prior is chosen for $\bz$, which is usually the multivariate unit
Gaussian distribution, i.e., $\cN (0, \bI)$. Then, $\bx (i)$ is a data
point drawn from the 
distribution $p(\bx|\bz) p(\bz)$, where $p(\bz)$ and $p(\bx|\bz)$ are
the \emph{a priori} distribution and likelihood
of the latent variables, respectively.
This posterior is usually assumed to be 
$\cN (\mu_{\theta(\bx)},\sigma^2_{\theta(\bx)} \bI)$, where 
$\mu_{\theta (\bx)}$ and $\sigma^2_{\theta(\bx)}$ are obtained by 
a multilayer neural network that is characterized by the network 
parameter set $\theta$ and called the decoder (in most cases, 
$\sigma^2_{\theta(\bx)}$ is assumed to be fixed).
The encoder, which is another network characterized by the network 
parameter set $\phi$, is used to map $\bx$ to $\bz$
by finding $q_\phi (\bz|\bx)$. With a given dataset,
the encoder and decoder are trained to minimize the reconstruction
error.

\subsection{VAE-based Anomaly Detection}

%Once the VAE is trained with a training dataset, the anomaly detection can be carried out. 
% In this subsection, we briefly present
% the anomaly detection and the application of VAE or VQ-VAE to the anomaly
% detection. 

Denote by
$f_0 (\bx)$ the distribution that generates
the training vectors, i.e., $\bx_{(i)} \sim f_0 (\bx)$.
In other words, $f_0 (\bx)$ is the ground truth law of normal behavior.
Then, the fo llowing two hypotheses can be considered:
\begin{align}
H_0: \  \by \sim f_0 (\bx) \ \mbox{versus} \ 
H_1: \  \by \sim f_1 (\bx),
    \label{EQ:HH}
\end{align}
where $f_1 (\bx) (\ne f_0(\bx))$ is an anomaly distribution.
As a default uninformative prior, a uniform distribution
can be used for $f_1 (\bx)$ \cite{Steinwart05}.
Then, with known $f_0 (\bx)$, a set of anomalies can be defined
as $\cA (\tau) = \{\bx \in \cX\,|\, f_0(\bx) \le \tau\}$ with a threshold $\tau \ge 0$. If a test vector 
$\by$ belongs to $\cA(\tau)$, it can be seen as an anomaly. 
From \eqref{EQ:HH}, there are two types of decision errors:
Type 1 (or false-alarm) error that results from choosing $H_1$ when a
test vector follows $f_0 (\bx)$; and
Type 2 (or miss) error that results from choosing $H_0$ when a
test vector follows~$f_1 (\bx)$.

If $f_0 (\bx)$ is not available, but a dataset,
machine learning approaches can be used for anomaly detection \cite{Ruff21}.
In particular, as in \cite{An15}, VAE can be used, as 
the output of the trained VAE
is expected to be close to an input that is drawn from $f_0(\bx)$. On the other hand,
if the input is an 
anomalous test vector, the reconstruction from the VAE may not be close
to the input.
Thus, the following test statistics 
can be used:
\be 
T = ||\by - \hat \by||^2 \overset{H_1}{\underset{H_0}{ \gtrless} } % \defhr 
\gamma, 
\ee 
where $\by$ and $\hat \by$
are the input and output of the trained VAE,
respectively, and $\gamma > 0$ is a decision threshold.

% \section{A Subspace-based Approach}
\section{HFR-AE: Algorithm and Design Principles}

This section delineates the process of the \emph{VAE-based HFR-AE framework (HFR-VAE)}, followed by presenting the rationale behind HFR-VAE through the lens of information~theory.

% This section describes HFR-AE utilizing projection onto an HFR subspace for anomaly detection. We also present the rationale behind HFR-AE through the lens of information~theory.

\subsection{Anomaly Detection via HFR-VAE}

Recall that $\bx_{(i)} \in \uR^L$ 
represents the $i$th training data to train the VAE.
Denote by
$\hat \bx_{(i)}$ the reconstruction
of the $i$th training data from the VAE.
The trained VAE is likely to yield a small 
% $\hat \bx_{(i)} \approx \bx_{(i)}$
% or the 
reconstruction error $\tilde \bx_{(i)}:= \hat \bx_{(i)} - \bx_{(i)}$. Since the dimension of the latent space is limited,
it is impossible (and to some extent undesirable) to make $\tilde \bx_{(i)}$ absolutely negligible, while
it could be possible to find a subspace where the reconstruction error is small enough. This subspace can characterize
the features of training vectors with reconstructions from
the trained VAE.

Suppose that the covariance matrix of $\tilde \bx_{(i)}$ is given by
\be 
\bC = \frac{1}{N} 
\sum_{i=1}^N \tilde \bx_{(i)} \tilde \bx_{(i)}^\rT,
\ee 
where $N$ is the number of the training vectors.
% and $\tilde \bx_{(i)} = \bx_{(i)} - \hat \bx_{(i)}$
% is the $i$th reconstruction error vector.
Let the eigendecomposition of $\bC$ be given by
\be 
\bC = \bE \bLambda \bE^\rT,
\ee 
where $\bE = [\bee_1 \ \ldots \ \bee_L]$
and $\bLambda = {\rm diag}(\lambda_1, \ldots, \lambda_L)$.
Here, $\lambda_l$ represents the $l$th smallest eigenvalue of
$\bC$ 
(i.e., $\lambda_1 \le \ldots \le \lambda_L$)
and $\bee_l$ is its corresponding eigenvector.
Clearly, we have
\be 
\lambda_l = \uE[|\bee_l^\rT \tilde \bx_{(i)} |^2],
\ee 
where the expectation is carried out over $i$.

Define
\be
\bE_\epsilon = [\bee_1 \ \ldots \ \bee_M],
\ee 
where
$ 
M = \max\{l : \ \lambda_l \le \epsilon\}. 
$
Here, $\epsilon \ll 1$.
Then, for any $i$,  we expect that
\be 
||\bE_\epsilon^\rT (\bx_{(i)}- \hat \bx_{(i)})||^2 \le M \epsilon 
\ee 
with high probability. 
This implies that  with
a sufficiently small $\epsilon$, the projection of the reconstruction error
onto the subspace of $\bee_1, \ldots, \bee_M$, i.e.,
${\rm Span}(\bee_1, 
\ldots, \bee_M)$, which
is referred to 
as the \emph{HFR subspace},
will be almost negligible.
In particular, the projection of $\bx \sim f_0(\bx)$
on to the HFR subspace,
i.e., $\bE_\epsilon^\rT \bx$, is to be reproduced with negligible errors.
This becomes a useful feature to characterize
the training vectors
as well as any test vectors that are drawn from the same distribution, 
$f_0(\bx)$.

If $\by$ is drawn from the same distribution
as the training vectors, $\bx_{(i)}$,
i.e., under hypothesis $H_0$, we can expect that
\be
||\bE_\epsilon^\rT (\by- \hat \by)||^2
\le M \epsilon 
\ee 
with a high probability. As a result, the following test
statistics can be considered for anomaly detection:
\be
T_{\rm sub} = ||\bE_\epsilon^\rT (\by- \hat \by)||^2
% \defhr 
\overset{H_1}{\underset{H_0}{ \gtrless} }
\gamma.
\ee

\subsection{An Information-Theoretic  Interpretation}

For an information-theoretic interpretation,
suppose that the reconstruction is given by
\be 
% \hat \bx = \bx + \tilde \bx,
\bx = \hat \bx  + \tilde \bx,
    \label{EQ:hxx}
\ee 
where $\tilde \bx \sim \cN(\b0, \bC)$ is the reconstruction error.
Once the VAE is trained, we can assume that
the reconstruction error, $\tilde \bx$, is uncorrelated
with the data sample, $\bx$. 
In this case,
if we assume that $\bx$ is a zero-mean Gaussian 
vector with covariance matrix $\bSigma$,
the mutual information between $\bx$ and $\hat \bx$ \cite{CoverBook} \cite{ChoiJBook2}
becomes
\begin{align}
{\sf I} (\bx; \hat \bx) 
 = \frac{1}{2} \log \det(\bSigma) \det(\bC^{-1}) .
\end{align}
Let $\sigma_l$ denote the $l$th eigenvalue of $\bSigma$.
Then, recalling
that the $\lambda_l$'s represent the eigenvalues of $\bC$,
the mutual information is
\begin{align}
 {\sf I} (\bx; \hat \bx)
 =  \frac{1}{2} \left(\sum_l \log \sigma_l - 
 \sum_l \log \lambda_l \right),
    \label{EQ:I2}
\end{align}
which shows that the mutual information increases as the
$\lambda_l$'s decrease.
From \eqref{EQ:hxx}, we can see that $\hat \bx$ and $\bx$ are the output and input of a certain MIMO channel, respectively, with the mutual information in \eqref{EQ:I2}. We can divide this channel into two channels to get a useful channel for anomaly detection.

We now decompose the signals by projecting them on to two orthogonal subspaces
as follows:
\begin{align}
\bv_1 & = \bE_\epsilon^\rT \bx, 
\quad  \hat \bv_1 = \bE_\epsilon^\rT \hat \bx 
= \bv_1 +\bE_\epsilon^\rT \tilde \bx \cr
\bv_2 & = \bE_+^\rT \bx, \quad \hat \bv_2  = \bE_+^\rT \hat \bx
= \bv_2 +\bE_+^\rT \tilde \bx,
    \label{EQ:Proj}
\end{align}
where $\bE_+ = [\bee_{M+1} \ \ldots \ \bee_L]$.
Let $\bSigma_1$ and $\bSigma_2$ be the 
covariance matrices of $\bv_1$ and $\bv_2$, respectively.
In addition, let $\sigma_{i,l}$ represent
the $l$th eigenvalue of $\bSigma_i$, $i \in \{1,2\}$.
Then, we can show that
\begin{align}
 {\sf I} (\bv_1; \hat \bv_1)
& =  \frac{1}{2} \left(\sum_{l=1}^M \log \sigma_{1,l} - 
 \sum_{l=1}^M \log \lambda_l \right) \cr
 {\sf I} (\bv_2; \hat \bv_2)
& =  \frac{1}{2} \left(\sum_{l=1}^{L-M} \log \sigma_{2,l} - 
 \sum_{l=M+1}^L \log \lambda_l \right),
    \label{EQ:Two_channel}
\end{align}
which are the mutual information of the following
two MIMO channels:
$\bv_1 \to \hat \bv_1$ and $\bv_2 \to \hat \bv_2$,
where the capacity of
the first channel is much higher than that of the second channel because $\lambda_l$, $l = 1,\ldots, M$, are less than or equal to $\epsilon \ll 1$.
For convenience, the first channel is referred to as the HFR channel
and the second channel the noisy 
or LFR.
Since the HFR channel is decided by the covariance matrix of the reconstruction
error or the trained VAE, it can be seen as a highly data-dependent channel, where the channel output is almost identical
to the channel input \emph{provided that the input
is drawn from the distribution of the training dataset, $\{\bx_{(i)}\}$}.
On the other hand, for a test data not drawn from the training dataset,
the channel output is not necessarily close to the channel input. As a result, the pair of the input and output of the HFR channel
can be used for anomaly detection. Note that the pair of the input and output of the LFR channel is not useful due to its too noisy channel output.

% However,
% the pair of the input and output of the LFR channel cannot
% be used for anomaly detection, because
% the channel output is noisy even if the input is drawn from the
% distribution of the training dataset, $\{\bx_{(i)}\}$.

% \begin{figure}[h]
% \begin{center}
% \includegraphics[width=\figwidth]{two_chns.png}
% \end{center}
% \caption{An illustration of the two channels induced
% by the training dataset.}
%         \label{Fig:two_chns}
% \end{figure}

%On the other hand, the second channel is not reliable. 

%can be seen 
%as a random projection 
%in terms of the data sample. Thus, the entropy 

 \begin{figure}[t]
  \includegraphics[width=\linewidth]{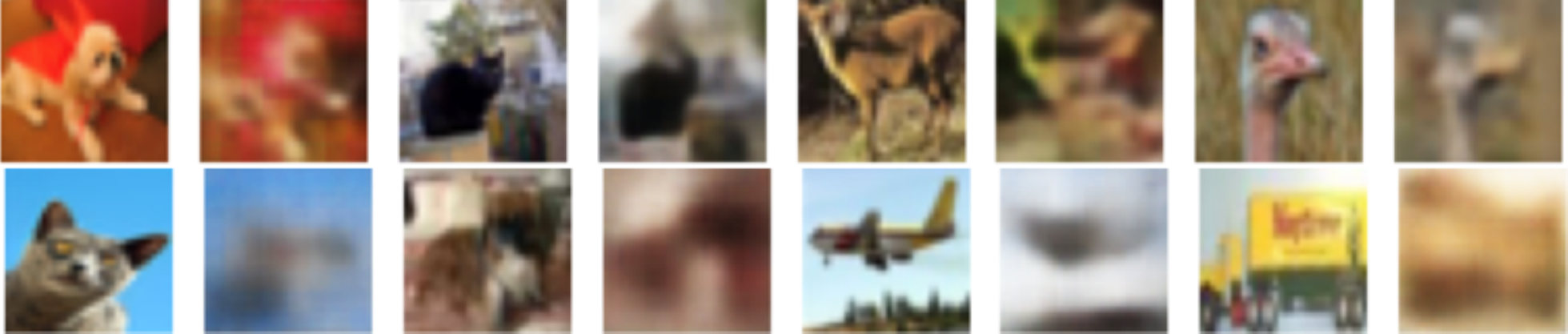}
  \caption{Reconstructed images by VQ-VAE on true and false datasets in row 1 and 2, respectively.}
  \label{fig:recons} 
  \vspace{-10pt}
\end{figure}

\begin{figure}
  \includegraphics[width=\linewidth, height=4.5cm]{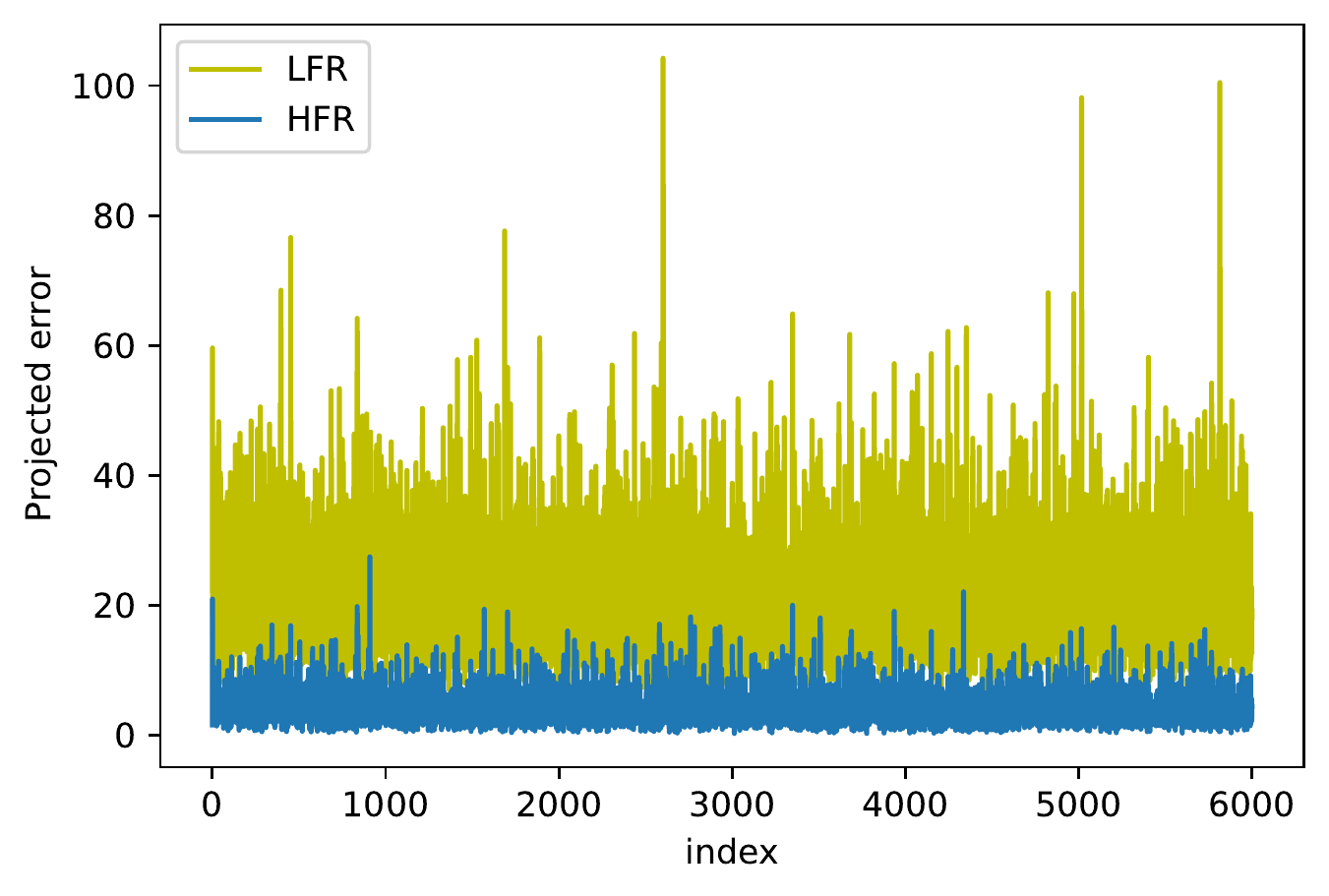}
  \caption{L2 norm of HFR/LFR subspace projected errrors.}
  \label{fig:vq_h_b} 
  \vspace{-10pt}
\end{figure}

\section{Experiments}

% \textbf{A. Experimental Setup} 

\textbf{Experimental Settings}. We consider VQ-VAE. VAE, and Vanilla AE architectures. For all models, the encoder consists of $2$ strided convolutional layers with stride $2$ and kernel size $3$x$3$, followed by two residual $3$x$3$ blocks each of which consists of a $3$x$3$ convolutional (Conv) layer and a $1$x$1$ Conv layer. All these layers have $256$ hidden units. The decoder has two residual $3$x$3$ blocks, followed by two transposed Conv layers with stride $2$ and window size $4$x$4$. Activation functions are rectified Linear Units (ReLU). For VQ-VAE, the discrete latent space is chosen as $8$x$8$ embedding space with $K=128$ quantization levels and $D=256$ dimension per quantized codeword. The commitment loss weight of VQ-VAE is $0.25$. To train these models, we use the ADAM optimizer with learning rate $2e-4$ and evaluate the performance after $100$ epochs with batch size $128$. We consider the CIFAR-10 dataset comprising 60k images of $32$x$32$x$3$ with 6k images of each class. We use 50k images from each of 6 classes to train the model on the right data as training set and total of 6k images as the test set. This test set has 5k images from the same 6 classes as the right data and 1k images from the remaining 4 classes as the false data, resulting in the reconstruction output as Fig.~\ref{fig:recons} visualizes. By default we consider VQ-VAE unless otherwise specified.
% We measure the performance of anomaly detection used AUROC as the anomaly detection fi.
    
\begin{table}
  \begin{center}
    \caption{Impact of HFR subspace threshold $\epsilon$ on the maximum eigenvalues of the subspace, and MSE of the projected errors with right and false data.}
    \label{tab:table3}
    \begin{tabular}{c|c|c|c}
      \toprule % <-- Toprule here
      \textbf{Threshold} & \textbf{Max}&\multicolumn{2}{c}{\textbf{MSE w. HFR-VAE}}
      \\\textbf{$\epsilon$}&\textbf{eigenval.}&\textbf{right data}&\textbf{false data}
      \\
      \midrule % <-- Midrule here
      0.00005 & 0.001462&0.7265 &0.8053\\
      0.0001 & 0.002924&1.310&1.4910\\
      0.0005 &  0.01462&4.599&5.377 \\
      0.001 & 0.02933&7.631&8.892\\
      0.0015 & 0.04396&10.44 &12.09\\
      \bottomrule % <-- Bottomrule here
    \end{tabular}
  \end{center}
\end{table}

\begin{table}
  \begin{center}
    \caption{Maximum eigenvalues of En vector obtained by eigendecomposition and AUROC by HFR-AE with varying bottleneck dimension.}
    \label{tab:table4}
    \resizebox{\columnwidth}{!}{\begin{tabular}{c|c|c|c|c|c|c|c}
    
      \toprule % <-- Toprule here
    \textbf{Latent}&
      \textbf{Max} 
  &\multicolumn{2}{c|}{\textbf{AUROC}}
  &\multicolumn{2}{c|}{\textbf{MSE w. VAE}}
  &\multicolumn{2}{c}{\textbf{MSE w. HFR-VAE}}\\
% \cline{3-6}
             \textbf{dim.}&\textbf{eigenval.}&\textbf{HFR-VAE}&\textbf{VAE} &\textbf{right}&\textbf{false}
             &\textbf{right}&\textbf{false}\\
    %   \textbf{bottleneck Dim.}&\textbf{Max En}&\textbf{HFR AUROC}&\textbf{Recons AUROC}&\textbf{right recons err.}&\textbf{false recons err.} \\
      \midrule % <-- Midrule here
      32 &   0.0953&0.584&0.515&0.063&0.063&14.81&16.27\\
      64 &  0.0176&0.595&0.560&0.011&0.012&5.24&6.28\\
      128 &  0.0149&0.595&0.563&0.0098&0.010&4.68&5.48\\
      256 & 0.0125&0.594&0.576&0.0082&0.0090&4.02&4.91\\
      512& 4.2e-05&0.594 &0.581&0.0077&0.0085&3.97&4.44\\
      1024&8.2e-06&0.593&0.588& 0.0070& 0.0080&3.67&4.12\\
      
      \bottomrule % <-- Bottomrule here
    \end{tabular}}
  \end{center}
  \vspace{-10pt}
\end{table}

\textbf{HFR vs. LFR Subspace Projected Errors}.
We use the eigendecomposition of the reconstruction error vector projected onto the HFR subspace i.e.,$\bE_\epsilon^\rT \bx$ having  $\epsilon = 0.001 \ll 1$. 
% The indices of eigenvalues lying below the threshold $\epsilon$ is chosen to create a new subspace. For comparing our hypothesis, we project our error vector to this HFR subspace and compare the Norm of error values to the original subspace comprising all eigenvalues.
With the test dataset for right samples, Fig.~\ref{fig:vq_h_b} reports the L2 norm of the reconstruction error in the subspace composed of large eigenvalues in orange (LFR subspace), the projected reconstruction error in the smaller eigenvalue subspace in blue (HFR subspace). It shows that the range of L2 norm error for the right data projected onto the HFR subspace is much lower with less variance than that under the LFR subspace. Such L2 norm of HFR-subspace projected right data will be distinctively distinguished from the L2 norm of HFR-subspace projected false data that are unlikely to be low.

  \textbf{Impact of HFR Subspace Threshold}. The HFR subspace threshold $\epsilon$ partitions the subspace made by eigenvalues, affecting the HFR subspace dimension and the projected error in that space. In Table~\ref{tab:table3}, we observe the trend of maximum eigenvalue increases with $\epsilon$. As the threshold increases, the reconstruction error, measured using mean squared error (MSE) between reconstructed and original images, also increases both on right instance as well as false instance. The MSE on false instance remains greater which leads to anomaly instances.
  As we further decrease the threshold, model reduces its efficiency to distinguish between normal instances and outliers, showing the existence of an optimal $\epsilon$. These thresholds also depends and changes its effectiveness on changing the size of bottleneck dimension.
  The given result is for latent dimension=$265$ in Tab.~\ref{tab:table3}. When we increase the dimension, the lowest reconstruction MSE comes around $\epsilon=0.0005$. Such an optimal $\epsilon$ can be found by simply feeding multiple samples, as opposed to existing IB-based AE frameworks that require re-training to optimize their bottleneck dimensions \cite{Kingma14}, quantization levels \cite{Oord17}, and loss regularization \cite{higgins2017beta}.

   \begin{figure}
  \includegraphics[width=\linewidth, height=4.5cm]{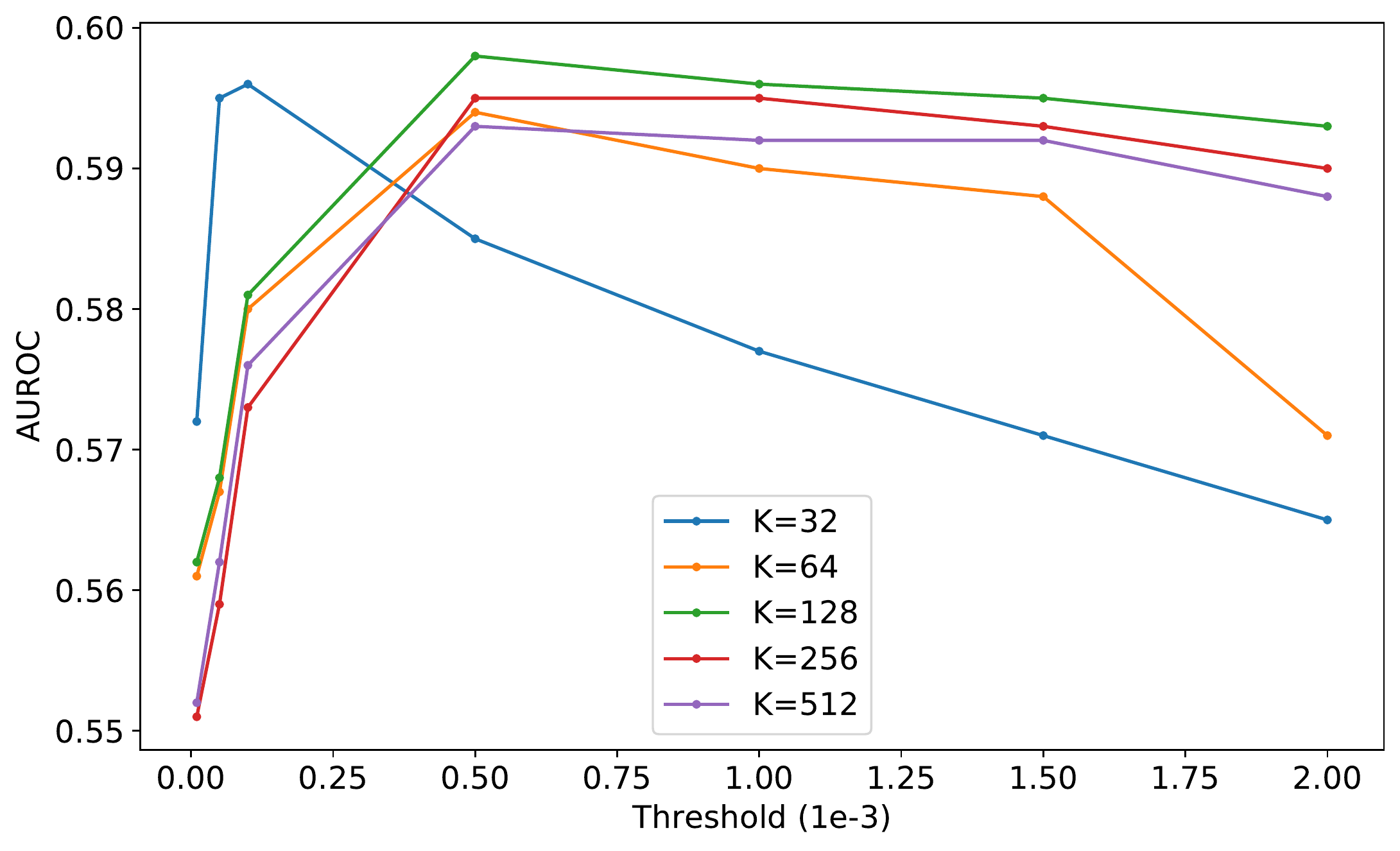}

  \caption{AUROC with respect to the HFR subspace threshold $\epsilon$ on the bottleneck dimension $K$.}
  \label{fig:vae_lambda}
  % \vspace{-10pt}
\end{figure}

\begin{table}[t]
  \begin{center}
    \caption{Mean and deviation of AUROC under different AE architectures.}
    \label{tab:table1}
    \begin{tabular}{c|c|c}
      \toprule % <-- Toprule here
      \textbf{Architecture} & \textbf{w.o. HFR-AE} & \textbf{w. HFR-AE}  \\
      \midrule % <-- Midrule here
      Vanilla AE & 0.569$\pm$0.03  & 0.593$\pm$0.01   \\
      VAE &0.551$\pm$0.02 &0.591$\pm$0.01  \\
      %NVAE &0.520\pm0.02 &0.678\pm0.01  &0.456\pm0.003 \\
      VQ-VAE & \textbf{0.573$\pm$0.0} & \textbf{0.598$\pm$0.03} \\
      
      \bottomrule % <-- Bottomrule here
    \end{tabular}
  \end{center}
  \vspace{-10pt}
\end{table}

% \begin{table}[h!]
%   \begin{center}
%     \caption{Mean and variance of HFR, LFR error with varing thresholds($\epsilon$) for selecting eigen values for HFR subspace}
%     \label{tab:table1}
%     \begin{tabular}{c|c|c|c|c}
%       \toprule % <-- Toprule here
%       \textbf{Thresholds} & \textbf{HFR mean} & \textbf{HFR var.} & \textbf{LFR mean}& \textbf{LFR var.}\\
%       \midrule % <-- Midrule here
%       0.00005 &  0.716 & 0.365 & 31.99 & 180.92  \\
%       0.0001 &  1.345 & 1.019 & 31.35 & 172.99   \\
%       0.0005 &  4.93 & 8.865 & 27.77 & 135.68   \\
%       0.001 & 8.144 & 20.868 & 24.55 & 108.721 \\
%       0.0015 & 11.11 & 35.332 & 21.568 & 88.86  \\
%       \bottomrule % <-- Bottomrule here
%     \end{tabular}
%   \end{center}
% \end{table}

% \begin{table}[h!]
%   \begin{center}
%     \caption{Mean and deviation of AUROC[\%] with varriying bottleneck dimension }
%     \label{tab:table1}
%     \begin{tabular}{c|c|c}
%       \toprule % <-- Toprule here
%       \textbf{bottleneck dimension} & \textbf{Reconstruction error} & \textbf{HFR error} & \\
%       \midrule % <-- Midrule here
%       32 & 0.526\pm0.03  & 0.585\pm0.01   \\
%       64 & 0.557\pm0.03  & 0.590\pm0.01   \\
%       128 &0.568\pm0.02 &0.598\pm0.01  \\
%       256 &0.573\pm0.02 &0.595\pm0.01  \\
%       512 & 0.562\pm0.02 & 0.592\pm0.03 \\
      
%       \bottomrule % <-- Bottomrule here
%     \end{tabular}
%   \end{center}
% \end{table}

\textbf{Impact of IB}. Next, we vary the bottleneck dimension of AE archtiectures, and observe the changes in accuracy on finding anomalies and the max eigenvalues of the HFR subspaces. As shown in Tab.~\ref{tab:table4}, with higher bottleneck dimension, more information can be stored at the bottleneck of the input image, thereby reducing the reconstruction errors. Meanwhile, the HFR subspace projected errors are convex shaped over the bottleneck dimension. Maximum accuracy can be achieved on the bottleneck dimension of 128. Consequently, Fig.~\ref{fig:vae_lambda} captures the variations in both $\epsilon$ and bottleneck dimension, showing that the highest AUROC can be achieved at the bottleneck dimension $128$ and $\epsilon=0.0005$.

  \textbf{Impact of AE Architectures}. Finally, to validate the feasibility of our HFR-AE framework under different AE architectures, in addition to HFR-VAE, we additionally consider the HFR-AE frameworks with Vanilla AE (HFR-Vanilla) and VQ-VAE (HFR-VQVAE). With the common bottleneck dimension $256$, Tab.~\ref{tab:table1} shows applying the HFR-AE framework improves AUROC under all considered architectures. The highest AUROC is achieved under the VQ-VAE architecture that also achieves the higest AUROC without HFR-AE.
  
  % that all these models have a similar encoder, decoder architecture and have been tested on a common bottleneck dimension of 256. It has better performance on all the tested Autoencoder models from the reconstruction based model having highest performance in the VQ-VAE model with vector quantized distribution in the latent space.
  
  % Tab.~\ref{tab:table1} shows the comparison of HFR subspace projecting method among different Autoencoder models. 

% \begin{figure}[h!]
%   \includegraphics[width=\linewidth]{Images/vae_threshold.pdf}
%   \caption{Variation of the cut-off Eigen value b/w En and ES vs threshold($\epsilon$) on varying bottleneck dimension }
%   \label{fig:vae_lambda}
% \end{figure}

% The obtained results suggest that our approach 
% outperforms a standard reconstruction error based method on VAE. Projecting the error to a reduced dimension choosing right features subspace, increases our probabilty to distinguish between a normal instance and an anomaly.
% We also observed that the method is sensitive to the pixel intensity of 
% the anomaly. Anomalies with intensities near the expected 
% intensities are often missed. This can be due to the anomaly 
% scored being calculated as the residual of pixel intensities. 

\section{Conclusion}
In this article we put forward to a novel AE-based anomaly detection framework, named HFR-AE, that projects inputs into a trained AE's HFR subspace so as to increase the output gaps between normal and anomalous samples. While improving AUROC for anomaly detection, HFR-AE is architecture-agnostic, and optimizing its key hyperparamter (i.e., HFR subspace threshold) is free from re-training, as evidenced by extensive simulations. To cope with dispersed training data in reality, extending this standalone HFR-AE framework to distributed HFR-AE frameworks by leveraging federated and other distributed learning methods \cite{park2021communication} could be an interesting topic for future research.

% Simulations corroborate that HFR-AE improves AUROC for anomaly detection under different AE architectures, 

% We presented a novel unsupervised anomaly detection method based on VAEs that has improved results over an existing standard VAE approach of using reconstruction error as an anomaly score. We formulate the large reconstruction error problem by projecting the vector onto a subspace of $\bee_1, \ldots, \bee_M$,which will be lower than the actual value. We introduced a hyperparameter for setting threshold ($\epsilon$) to control the selection of eigen docomposed values to construct High fidility Subspace. So this can be used as an important feature to characterize anomalies. With
% an appropriately tuned hyperparameter, we show that our
% learned representations improve the mean average precision
% of our anomalous data detection task.

\bibliographystyle{ieeetr}
\bibliography{references}
\end{document}